\documentclass{article}

     \usepackage[final]{neurips_2021}

\usepackage[utf8]{inputenc} 
\usepackage[T1]{fontenc}    
\usepackage{hyperref}       
\usepackage{url}            
\usepackage{booktabs}       
\usepackage{amsfonts}       
\usepackage{nicefrac}       
\usepackage{microtype}      
\usepackage{xcolor}         
\usepackage{graphicx}       
\usepackage{amsmath}        

\title{An Emulation Framework for Fire Front Spread}

\author
  {
  Andrew Bolt, Joel Janek Dabrowski, Carolyn Huston, Petra Kuhnert \\
  ~\\
  Data61 / CSIRO, Australia
}

\hypersetup{pdfborder={0 0 0},colorlinks=true,urlcolor=blue,linkcolor=blue,citecolor=blue,bookmarks=true}
\hypersetup{pdftitle={An Emulation Framework for Fire Front Spread}}
\hypersetup{pdfauthor={Andrew Bolt, Joel Janek Dabrowski, Carolyn Huston, Petra Kuhnert}}

\begin{document}

\maketitle

\begin{abstract}
  Forecasting bushfire spread is an important element in fire prevention and response efforts. Empirical observations of bushfire spread can be used to estimate fire response under certain conditions. These observations form rate-of-spread models, which can be used to generate simulations.  We use machine learning to drive the emulation approach for bushfires and show that emulation has the capacity to closely reproduce simulated fire-front data. We present a preliminary emulator approach with the capacity for fast emulation of complex simulations. Large numbers of predictions can then be generated as part of ensemble estimation techniques, which provide more robust and reliable forecasts of stochastic systems.
  
  
\end{abstract}


\section{Introduction}

Bushfires pose a serious threat to communities and natural flora and fauna across wide regions of Australia, as well as internationally. Simulated fires provide valuable data for first responders to assess vulnerable areas and the risk of a firefront impacting communities as well as being able to formulate response strategies for threatening fires. Simulation platforms such as Spark~\citep{Spark2015} and Phoenix~\citep{Tolhurst2010} use various techniques to predict how a fire front will progress through time. Underpinning such simulations are empirical rate-of-spread (ROS) calculations. These calculations determine how quickly a fire burns given a fuel source and varying environmental conditions such as temperature, slope, wind speed and wind direction~\citep{cruz2015guide}.

A single simulation instance generates one possible future fire front. To generate uncertainty estimates of a fire reaching a given location requires running ensembles of simulations under various environmental conditions. Generating large ensembles becomes computationally taxing and may be a prohibitive barrier to this type of analysis.

Emulation using machine learning is a method that attempts to mimic a slow running and highly parameterized process model using training examples. We develop an emulator that approximates a simulated fire front and discuss how surrogate models of this type could be used more efficiently in the future to characterise a broad range of fire scenarios simulated from varying environmental setups..

\section{Modelling}
\label{sec:modelling}

\subsection{Data}
\label{sec:data}

We use a data set of 200 simulated fires, generated using the Spark platform under real world meteorology and land input conditions in Australia. These trials are a subset of SPARK runs conducted as part of CSIRO's commercial work in the bushfire space.  We aim to closely reproduce the simulated data using an emulator that requires a fraction of the computational resources. This makes running large ensembles to explore a broader range of fire scenarios, a more feasible proposition.


Input data for the simulation consists of a topographical map\footnote[1]{Topography data sets derived from Geoscience Australia SRTM-derived 1 Second Digital Elevation Models Version 1.0. Data is publicly available under Creative Commons Attribution 4.0 International Licence.}, weather data\footnote[2]{Meteorological time series data sets derived from Bureau of Meteorology automated weather station data .}, and land classification map\footnote[3]{Land classification data sets  derived from Department of Agriculture and Water Resources (ABARES) Land Use of Australia 2010-11 data set. Data is publicly available under Creative Commons Attribution 3.0 Australia Licence.}. The resolution of spatial data is 30mx30m for each pixel. Weather station data is polled every 30 minutes.

Key pre-processing steps in our machine learning pipeline include: converting heightmaps to $x$ and $y$ gradient components using a Sobel edge algorithm; converting wind speed and direction to $x$ and $y$ components; expressing distances in pixel (30m) units and times in interval (30 minutes) units; and creating training samples by cropping simulated images to $256\times256$ pixel squares, centred about an active fire source. Cropping is used because the memory requirements for training neural networks using large images is prohibitive.

\subsection{Neural Network Architecture}
\label{sec:architecture}

We approach the emulation of fire spread using a neural network (NN) framework due to their versatility as well as their success in structuring emulators for other problems~\citep{sekou2019,wang2019,allaire2021,burge2020}. The design of the NN should address some challenging aspects of the input feature and output space, such as image size, speed to generate outputs and a NN architecture that can cope with a mix of spatio-temporal inputs. 

In terms of the size of the images that convey the fire front, each sample is allowed to vary, therefore accepting inputs to be of varying size. Fully convolutional networks are able to handle a variable input array size, so this is a natural choice.

A focus for emulation is speed and to ensure that the overall complexity of the network is minimized while still maintaining performance~\citep{Thiagarajan2020, kasim2020}. Our approach is to downscale the spatial data (topography and land/fuel type) through strided convolution operators. This smaller set of latent features are then updated by each time interval until a final state is reached. This is far less computationally expensive than applying updates directly to a full sized array. The final latent state is then upscaled by transposed strided convolution operators where a final fire shape is output.

Finally, incorporating weather data is itself difficult since it more closely resembles time series data than image data. We could simply treat each data point as a uniform array of values and approach the problem using standard convolutional techniques. Undesirably, this greatly increases the number of convolutional operations that must be performed, which taxes memory and processing power. Instead, we transform the latent terrain layers so their depth dimension matches that of the weather input. By multiplication we transform these layers into an input with the correct spatial dimensions. This process is repeated for each new weather input until the final fire shape is produced. 

\begin{figure}[!b]
    \centering
    \includegraphics[width=1.0\textwidth]{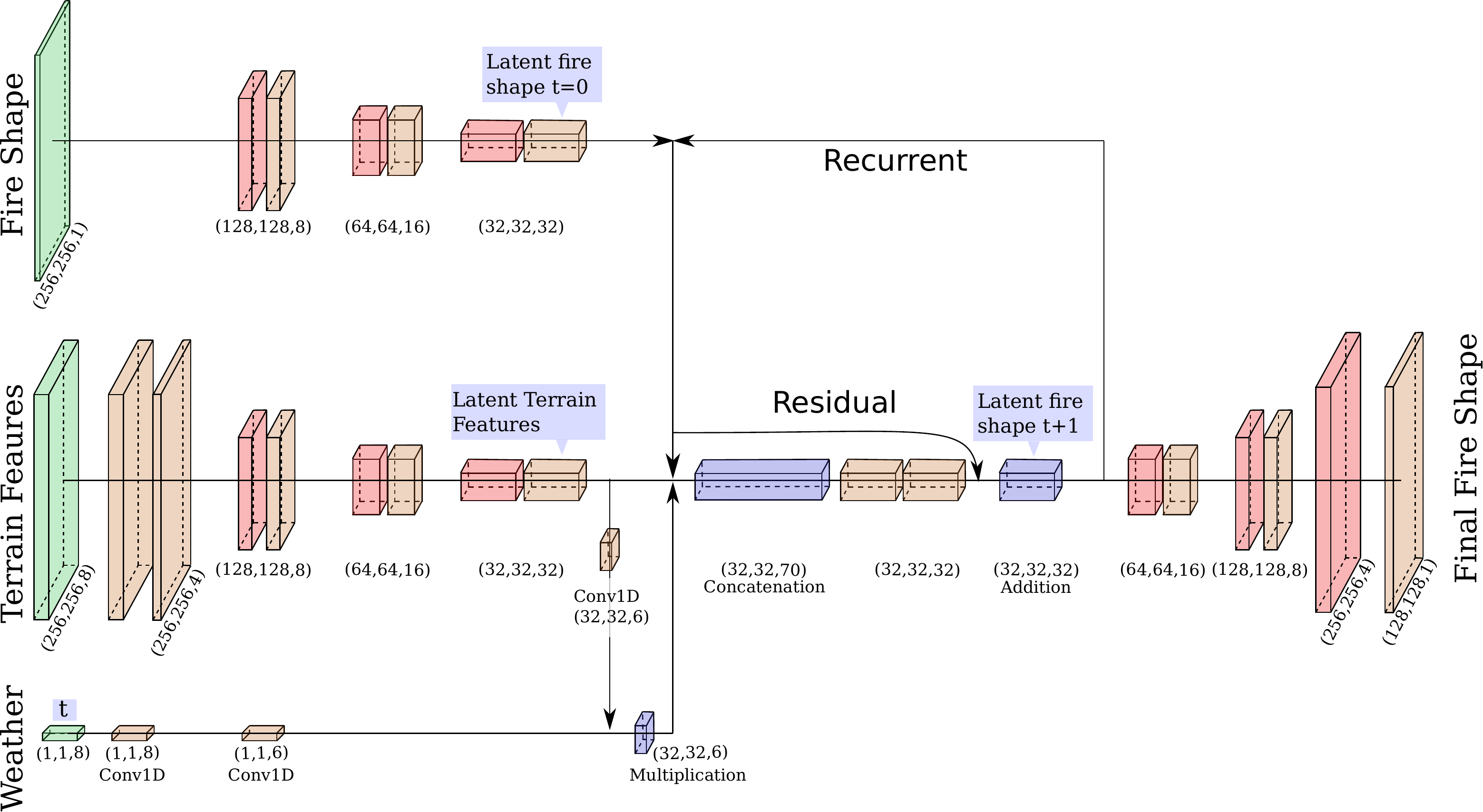}
    \caption{\small The emulator architecture. Red blocks represent strided convolutional (or transposed convolutional) layers with kernel size 4 and stride size 2. Orange blocks represent convolutional layers with kernel size 3 and stride size 1. The fire shape and terrain inputs are 2D arrays. In training we use 256$\times$256 pixels, though the model is fully convectional and can handle arbitrarily sized arrays. Weather inputs are a set of scalar values (eg.~temperature, wind speed). These values are updated in successive steps as part of the input to the recurrent component of the model. The state of this recurrent cell are the latent fire shape features. Once the final weather input is processed upscaling is used to restore the final predicted fire shape.}
    \label{fig:architecture}
\end{figure}

Figure~\ref{fig:architecture} shows a schematic for the neural network we deploy. A sample input image size of $256\times 256$ pixels is used as a demonstration. The model uses a total of 106,532 trainable parameters. Note that there are only 21,248 trainable parameters in the residual block. The final model output layer is a spatial layer with values corresponding to arrival times of the fire.

\subsection{Training the Neural Network}


As the size of the images supplied for each fire simulation vary, some being quite large (up to 2048 pixels on an edge), we take the approach of cropping the images to 256 by 256 pixels in order to reduce the memory requirements needed for training. The cropping is centered around an actively burning region on the perimeter, and random rotation and flipping is performed.


The loss function, $\mathcal{L}$ of an image, $\mathcal{P}$ is represented as 
\begin{align*}
    \mathrm{\mathcal{L}(\mathcal{P})} &= \mathrm{log}_{10}\left(\frac{MSE_{o} + \tau}{MSE_p + \tau} \right) \\
    \label{eq:loss_function}
\end{align*}
where $MSE_o = \frac{1}{n} \sum_{i=1}^n (y^{(o)}_i-y_i)^2$, $MSE_p = \frac{1}{n} \sum_{i=1}^{n} (y^{(p)}_i-y_i)^2$ and $y_i^{(o)}$ is the initial observed fire arrival time at the $i$-th pixel, $y_i$ is the target future arrival time at the $i$-th pixel, and $y_i{(p)}$ is the predicted future fire arrival time at the $i$-th pixel. The term $\mathrm{\tau}$ is a very small positive number to avoid asymptotic instability. The loss function is the log ratio of the mean squared error (MSE) of the initial and final arrival maps over the predicted and final arrival maps. This can be thought of as the improvement of the emulator over simply "doing nothing". The loss function was chosen since it does not weight samples with fast fire growth and more strongly than samples with limited fire growth.

The model is implemented using TensorFlow 2.0 and trained for 400 epochs. We used the ADAM optimizer \citep{Diederik2015} and a batch size of 16. We withhold a test set with a split of 0.2. This set remains un-cropped.


\section{Results}
\label{sec:results}

\begin{figure}
\centering
\begin{minipage}{.50\textwidth}
  \centering
  \includegraphics[width=.95\linewidth]{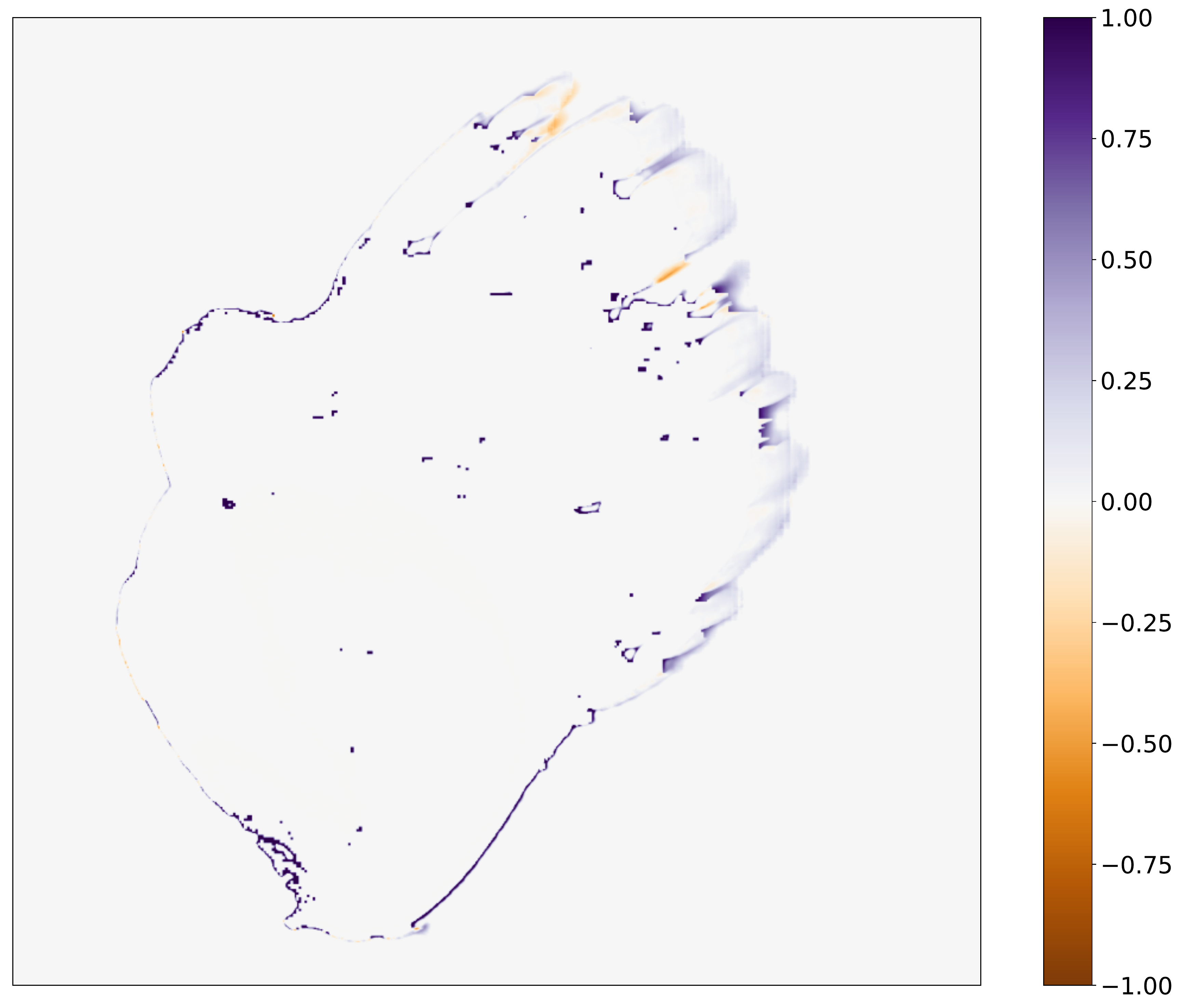}
  \caption{\small The difference between predicted and comparison (simulated) fire arrival times for a single interval (30 minutes) on a test sample. The interval is normalised so 0 represents the beginning and 1 the final arrival time. Positive values (purple) represent false-positives and negative values (orange) represent false-negatives.}
  \label{fig:test_prediction}
\end{minipage}%
\hspace{0.06\textwidth}
\begin{minipage}{.43\textwidth}
  \centering
  \includegraphics[width=.95\linewidth]{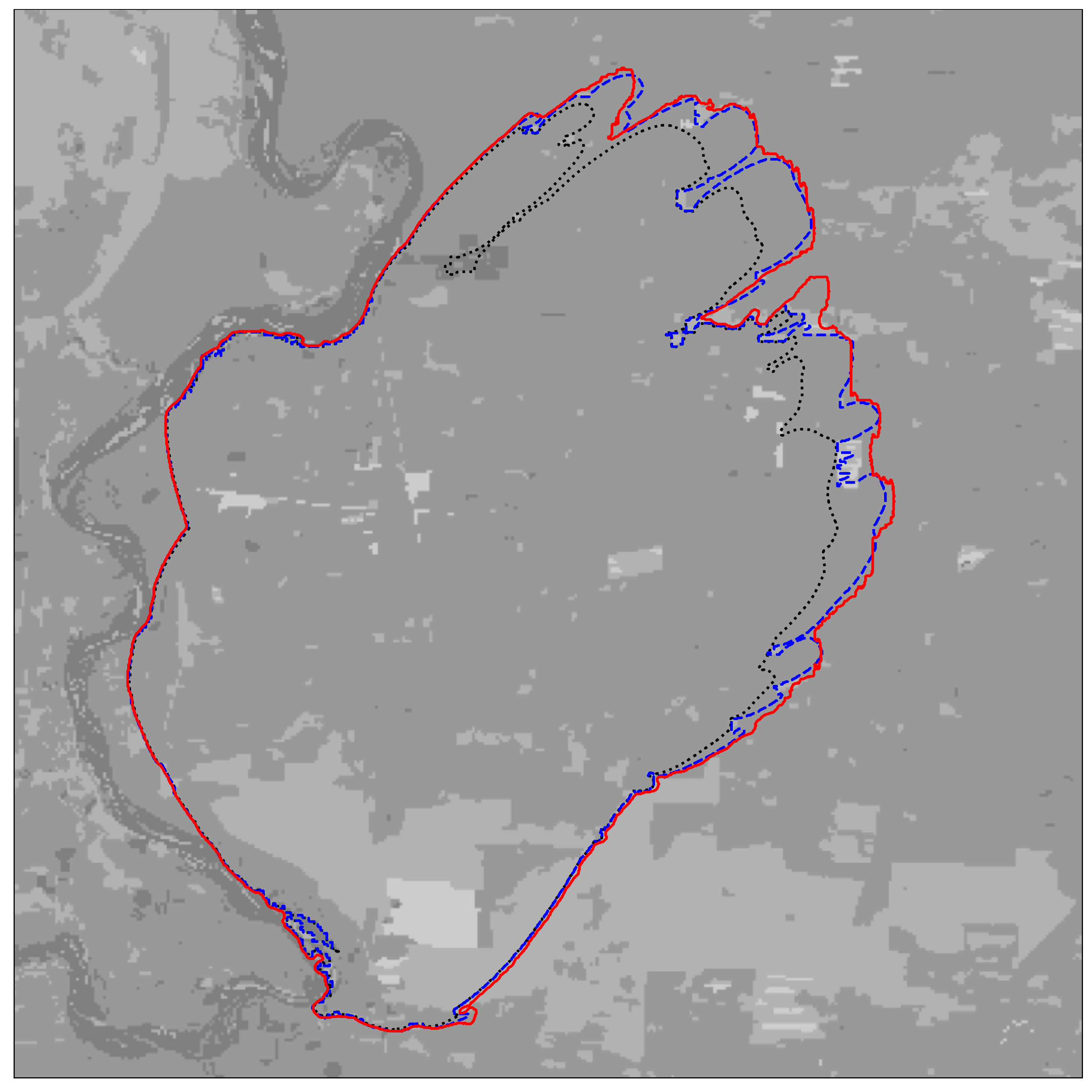}
  \caption{\small The initial fire front (dotted, black) shown against the predicted fire perimeter (solid, red) and the comparison fire perimeter (dashed, blue) for the same sample in Figure~\ref{fig:test_prediction}. These perimeters are overlaid on the various land classes used by the model, shown in gray-scale.}
  \label{fig:test_contours}
\end{minipage}
\end{figure}

We present the model evaluation metrics in Table~\ref{tab:metrics}. There is a close agreement between results in the training and test sets, indicating that the model is generalising well. An additional benefit of cropping the training set is that it reduces the chance of over-fitting, and closely resembles principles from few-shot learning~\citep{wang2020}.

Figure~\ref{fig:test_prediction} displays predictions for a test sample simulated over a duration of 30 minutes. In this sample there is overestimation of the fires spread, in which the emulator estimates the fire spreading faster than the simulation. 

Figure~\ref{fig:test_contours} displays the fire front for this sample. Between the predicted and comparison (simulated) perimeters there is a good agreement of general shape. Broad features are in agreement, while intricate and narrow features present in the comparison perimeter are lost in the prediction. Importantly we see that the emulated behavior with respect to nonburnable terrain (darkest background shade) is consistent with the simulation. In particular we note that the fire does not advance North Easterly at the bend in the river on the left flank of the fire.

The recurrent component of our neural network allows for a series of intervals to build out a longer duration prediction. This is likely to be a better gauge of the emulator's performance and is an extension that we are currently exploring.

\begin{table}
    \label{tab:metrics}
    \centering
    \begin{tabular}{llll}\toprule
            Set & Loss  & Jaccard Score & Dice Score    \\\midrule
            training & -0.45 & 0.68          & 0.81       \\
            test     & -0.49 & 0.67          & 0.79 \\\bottomrule
    \end{tabular}
    \vspace{1ex}
    \caption{Model loss and evaluation metrics. Aggregate over all samples.}
\end{table}

In terms of bench-marking speed and memory requirements against conventional simulations we are still awaiting a more sophisticated analysis. Preliminary trials show that there is a speedup of around a factor of four. This may improve as the model architecture is refined, and the implementation is improved.

\section{Conclusion}
\label{sec:conclusion}

In this paper we have shown how a neural net can be constructed to efficiently emulate a spatio-temporal spread model. In this case we specifically focus on the emulation of fire front spread. While this work represents a preliminary investigation we show a respectable match between emulated and simulated results. If development of the emulator leads to a much faster representation of the physical process then this opens up a number of possible applications.


Of immediate interest is the use of emulators in ensemble forecasting and the generation of confidence intervals for fire front predictions. This approach allows the estimation of the likelihood of a fire reaching an area, rather than simply calculating the most likely fire front. Another area of interest is transfer learning to fine tune the model with the use of real fire examples. This could in principle lead to a neural network model that is more accurate than the original model and simulations that the emulator was developed on. 



\section*{Broader Impact}

In this paper we propose an architecture for emulating fire simulations from SPARK. The downscaling layers act to compress the data. This allows for a speedup over simulated fires which act on the full uncompressed topographical scale. The reduced memory requirements may also be useful since large scale SPARK simulations are often constrained by memory.

We have shown a modest reduction in processing time between our emulations and SPARK simulations. It is likely that these gains will improve as the code and architecture are further refined.

While we have demonstrated that emulation shows promise as a methodology for mimicking fire spread, this geo-spatial approach may have impact in a variety of similar problem spaces. Areas such as disease spread, pollutant spread, and pest spread all represent similar problem scopes where this emulation approach may be viable.

\bibliographystyle{unsrtnat}
\small
\bibliography{references}  


\end{document}